\documentclass[conference]{IEEEtran}
\IEEEoverridecommandlockouts
\usepackage{cite}
\usepackage{amsmath,amssymb,amsfonts}
\usepackage{algorithmic}
\usepackage{graphicx}
\usepackage{textcomp}
\usepackage{booktabs}
\usepackage{multirow}
\usepackage[colorlinks=true, linkcolor=red, citecolor=green]{hyperref}
\renewcommand{\figurename}{Figure}
\renewcommand{\tablename}{Table}
\usepackage[table]{xcolor}
\begin{document}

\title{LMDepth: Lightweight Mamba-based Monocular Depth Estimation for Real-World Deployment\\

\author{\IEEEauthorblockN{1\textsuperscript{st} Jiahuan Long}
\IEEEauthorblockA{
\textit{Shanghai Jiao Tong University} }
\and
\IEEEauthorblockN{2\textsuperscript{nd} Xin Zhou}
\IEEEauthorblockA{
\textit{University of Electronic Science and Technology of China}}
}


}

\maketitle

\begin{abstract}
Monocular depth estimation provides an additional depth dimension to RGB images, making it widely applicable in various fields such as virtual reality, autonomous driving and robotic navigation. 
However, existing depth estimation algorithms often struggle to effectively balance performance and computational efficiency, which poses challenges for deployment on resource-constrained devices.
To address this, we propose LMDepth, a lightweight Mamba-based monocular depth estimation network, designed to reconstruct high-precision depth information while maintaining low computational overhead. 
Specifically, we propose a modified pyramid spatial pooling module that serves as a multi-scale feature aggregator and context extractor, ensuring global spatial information for accurate depth estimation.
Moreover, we integrate multiple depth Mamba blocks into the decoder. Designed with linear computations, the Mamba Blocks enable LMDepth to efficiently decode depth information from global features, providing a lightweight alternative to Transformer-based architectures that depend on complex attention mechanisms.
Extensive experiments on the NYUDv2 and KITTI datasets demonstrate the effectiveness of our proposed LMDepth. Compared to previous lightweight depth estimation methods, LMDepth achieves higher performance with fewer parameters and lower computational complexity (measured by GFLOPs). We further deploy LMDepth on an embedded platform with INT8 quantization, validating its practicality for real-world edge applications.


\end{abstract}


\section{Introduction}


Monocular Depth Estimation  (MDE) involves predicting depth maps from single RGB images. Such depth information is the foundation of robotics and autonomous systems, as it enables machines to perceive and interact with their surroundings more effectively. For instance, in Simultaneous Localization and Mapping(SLAM)~\cite{sandstrom2024splat, PseudoRGBD}, MDE-derived depth maps facilitate the construction of precise 3D maps while simultaneously tracking the system’s location, which is essential for complex navigating environments. Similarly, in autonomous driving~\cite{zheng2024physical, schon2021mgnet, mancini2018j}, depth estimation is critical for understanding the spatial layout of the scene, enabling safe navigation, obstacle avoidance, and dynamic path planning. These applications highlight the importance of MDE in improving spatial awareness and decision-making in intelligent systems.

\begin{figure}
\centering
\includegraphics[width=1 \linewidth]{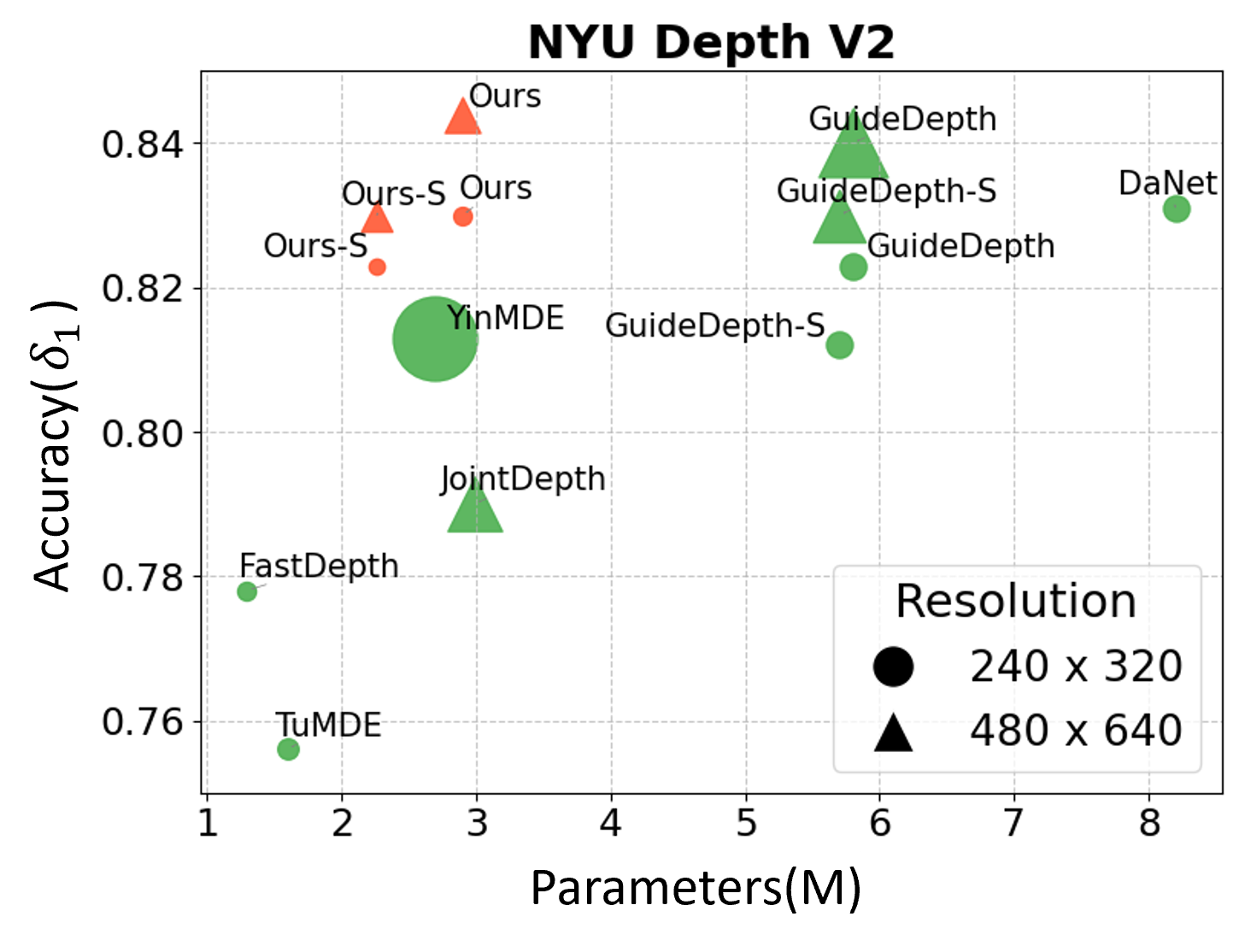}
\caption{\textbf{Complextity and performance comparisons of representative lightweight depth estimation methods.} The size of $\circ$ and $\Delta$ represents GFlops, with larger sizes indicating higher GFlops. Compared with other methods, our method has higher accuracy while keeping fewer parameters and lower GFlops across different image resolutions. }
\label{fig:intro}
\end{figure}

Numerous MDE methods have been proposed based on convolutional neural networks (CNNs)~\cite{laina2016deeper},~\cite{lasinger2019towards},~\cite{song2021monocular} and transformers~\cite{AdaBins},~\cite{newcrfs},~\cite{iebins}, achieving remarkable results in this field. However, most existing approaches primarily focus on improving accuracy while often overlooking the challenge of deploying on resource-constrained devices, such as unmanned aerial vehicles with limited storage and computational capabilities. To address this, current lightweight MDE methods~\cite{fastdepth, tumde, guidedepth} predominantly adopt CNN-based architectures to reduce computational complexity.  However, since the fixed size of receptive fields in CNNs, the model is more likely to be trapped in the local optimum during the optimization. In contrast, transformer-based architectures utilize global attention mechanisms to capture a broader receptive field, addressing the limitations of CNNs in modeling global context. However, they suffer from quadratic computational costs, which pose significant challenges for lightweight implementation.
Therefore, it is crucial to explore more efficient lightweight network architecture to balance the performance and computational overhead. The computational workflows of CNNs and transformers are briefly illustrated in Figures 2(a) and 2(b), respectively.

Recently, Mamba-based network architecture has significantly advanced the research of state space models on various vision tasks, including image classification~\cite{vim},~\cite{mambavision}, detection and segmentation~\cite{remamber},~\cite{U-mamba}. As an emerging framework, Mamba incorporates two key improvements in SSM: Firstly, MAMBA incorporates an input-dependent mechanism that dynamically adjusts the parameters of the State Space Model (SSM). Secondly, it adopts a hardware-aware design that processes data linearly with respect to sequence length, significantly boosting the computational efficiency on modern hardware systems. As shown in Figure 2(c), this mechanism uses an input-dependent matrix to selectively process the input. It can amplify or reduce certain input features based on their importance. The processed input is then projected into the state space, where it is used to update the state tokens effectively. Overall, Mamba's ability to combine efficiency and high performance makes it a compelling choice for advancing vision-related tasks.

In this paper, we propose LMDepth, a lightweight Mamba-based framework for Monocular Depth Estimation (MDE). It can effectively reconstruct depth information from single RGB image with low computational cost. Specifically,  we design a modified pyramid spatial pooling (MPSP) module that incorporates classification-aware strategy to generate depth classification bins. By leveraging these classification bins, LMDepth adapts effectively to diverse depth estimation scenarios. Additionally, we propose the composition of multiple Depth Mamba Blocks (DMBs) as the image decoder, enabling the efficient fusion of depth and image features through linear computations. By performing element-wise multiplication between the probability map and the bins, LMDepth predicts a high-quality depth while keeping a low computational overhead. As shown in Figure 1, our LMDepth and LMDepth-S achieve higher $\delta_1$ with fewer parameters and lower GFlops among lightweight methods on the NYUDv2 dataset, demonstrating its remarkable tradeoff of efficiency and performance in MDE task. 
The contributions of this work are summarized as follows:

\begin{itemize}
\item 
To the best of our knowledge, this is the first lightweight Mamba-based monocular depth estimation network, achieving high-precision depth estimation while keeping low computational overhead.
\item 
We propose a modified pyramid spatial pooling (MPSP) module that ensures global spatial features from RGB images. Compared to transformer-based decoder, we leverage multiple depth Mamba block (DMB) to efficiently decode the depth information from these global features.

\item 
Experiments on NYUDv2, iBims-1 and KITTI datasets demonstrate that our LMDepth outperforms representative lightweight methods by higher performance, fewer parameters and lower GFlops. 

\item We deploy LMDepth on a lightweight embedded platform and further validate it using real-world acquired data, demonstrating high inference efficiency, significant model size reduction, and strong practicality for real-world embedded deployment.

\end{itemize}

\begin{figure}
\centering
\includegraphics[width=1 \linewidth]{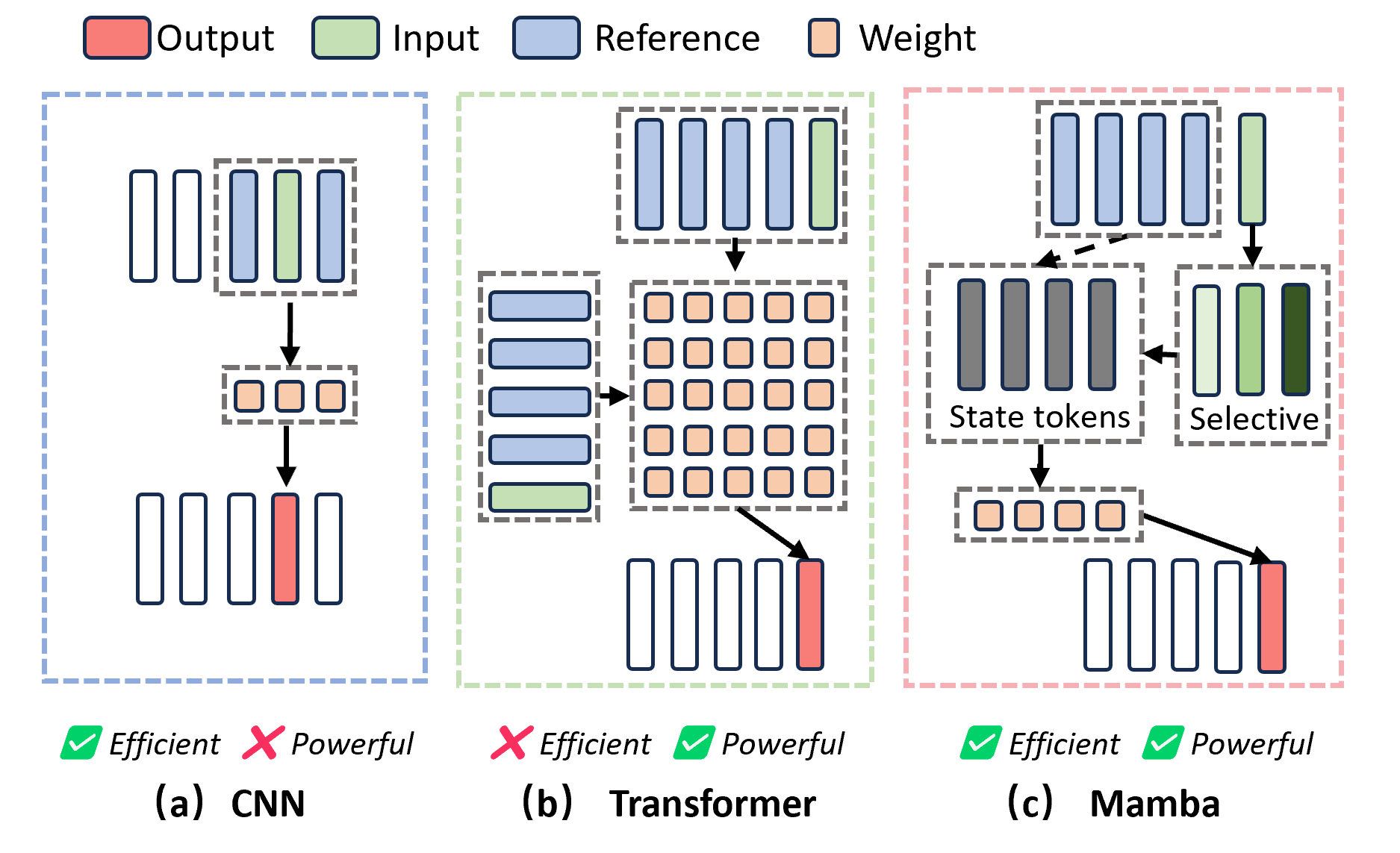}
\caption{\textbf{Comparison of computational processes of tokens among CNN, Transformer, and MAMBA.} 
It highlights why our Mamba-based depth estimation method is both efficient and powerful. 
Compared to CNN, MAMBA achieves an improved receptive field. In contrast to Transformer, MAMBA reduces computational complexity from quadratic to linear.
}
\label{fig:motivation`}
\end{figure}

\section{Related Work}
\subsection{Lightweight Monocular Depth Estimation}

Early monocular depth estimation methods relied on handcrafted features~\cite{make3d},~\cite{liu2014discrete}, but their limited representation ability led to poor accuracy. With the rise of deep neural networks, CNN-based methods significantly advanced this field. Eigen \emph{et al.}~\cite{eigen2014},~\cite{eigen2015} introduced a coarse-to-fine CNN framework, while Laina \emph{et al.}~\cite{laina2016deeper} improved accuracy with a fully convolutional encoder-decoder structure. Song \emph{et al.}~\cite{song2021monocular} further enhanced performance using a Laplacian pyramid-based decoder for combining hierarchical information.

Recently, self-attention-based models have emerged as strong alternatives. Ranftl \emph{et al.}~\cite{ranftl2020} demonstrated the superiority of Vision Transformers~\cite{vit} over CNNs with sufficient data. AdaBins~\cite{AdaBins} improved generalization by dynamically generating depth bins with transformers, while IEBins~\cite{iebins} refined depth predictions using a multi-stage search process. These advancements highlight the growing effectiveness of transformer-based models in improving monocular depth estimation accuracy and robustness. However, the quadratic computational complexity makes it very challenging to deploy on platforms with limited computational resources.

\begin{figure*}[t]  
\centering
\includegraphics[width=1\linewidth]{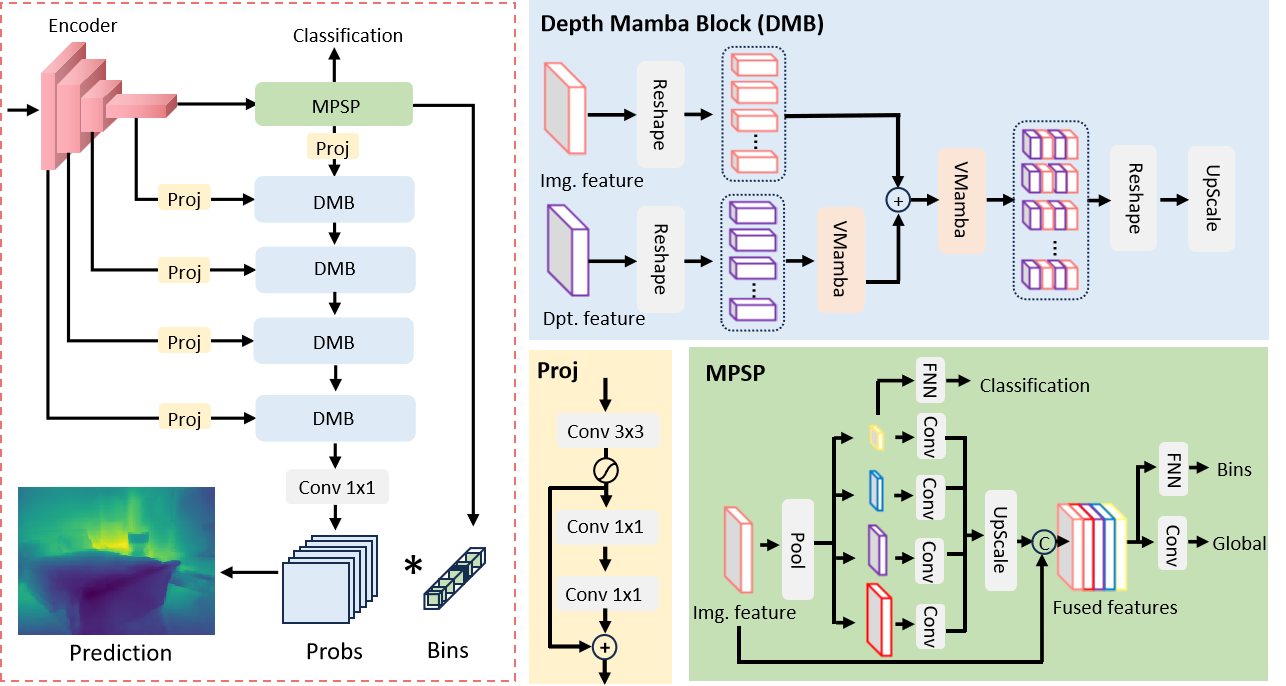}
\caption{\textbf{The overall network architecture of our LMdepth.} It consists of an image encoder, an MPSP head, and a decoder. The MPSP head outputs predicted scene classification and depth bins, while the decoder, composed of four custom-designed DMBs, predicts the depth distribution probabilities.}
\label{Fig:framework}
\end{figure*}

\subsection{State Space Models and Mamba}

State Space Models (SSMs), originally from control theory, have been adopted in deep learning for modeling long-range dependencies. Early works like LSSL~\cite{LSSL} and S4~\cite{s4} introduced efficient state representations and linear scalability with sequence length, overcoming limitations of CNNs and transformers. Later advancements such as S5~\cite{s5} and H3~\cite{h3} improved computational efficiency and achieved competitive performance in tasks like language modeling.

Recently, Mamba~\cite{mamba} emerged as a major breakthrough, providing linear-time inference and efficient training through a selection mechanism and hardware-aware optimizations. Mamba excels in tasks like medical image segmentation\cite{segmamba},\cite{segvm} and image dehazing\cite{haze}, demonstrating strong adaptability and efficiency. Given Mamba's balance between performance and efficiency, we hypothesize it can be applied effectively to lightweight monocular depth estimation. This task requires models with high accuracy and low computational overhead, where CNNs struggle with fixed receptive fields and transformers face quadratic complexity. 

In this work, we integrate Mamba into a lightweight depth estimation network to balance accuracy and efficiency. By leveraging its computational advantages and long-range modeling, we aim to enable high-performance depth estimation with reduced parameters and FLOPs, showcasing Mamba’s potential in resource-constrained vision tasks and establishing a foundation for SSM-based methods in efficient computer vision.

\section{methodology}
This paper aims to develop a lightweight Mamba-based architecture for the lightweight depth estimation task. We begin by introducing the foundational concepts of Mamba, followed by a detailed presentation of LMDepth, including its overall architecture, the Depth Mamba Block (DMB), the Modified Pyramid Spatial Pooling (MPSP), and the loss function design.

\subsection{Preliminaries}
SSM-based models, such as structured state space sequence models (S4) and Mamba, are inspired by continuous systems and are designed to capture the relationship between two functions or sequences, represented as $x(t) \in \mathbb{R} \mapsto y(t) \in \mathbb{R}$, through a hidden state $h(t) \in \mathbb{R}^N$. 

The evolution of the hidden state is governed by parameters $\mathbf{A} \in \mathbb{R}^{N \times N}$ for state transition, and $\mathbf{B} \in \mathbb{R}^N$, $\mathbf{C} \in \mathbb{R}^N$ for input and output projections, respectively. The continuous system is defined as:
\begin{equation}
\begin{aligned}
h'(t) &= \mathbf{A} h(t) + \mathbf{B} x(t), \\ 
y(t) &= \mathbf{C} h(t).
\end{aligned}
\end{equation}

S4 and Mamba are discrete versions of this system, using a timescale parameter $\Delta$ to transform the continuous parameters $\mathbf{A}$ and $\mathbf{B}$ into discrete ones, $\overline{\mathbf{A}}$ and $\overline{\mathbf{B}}$, via zero-order hold (ZOH):
\begin{equation}
\begin{aligned}
\overline{\mathbf{A}} &= \exp(\Delta \mathbf{A}), \\
\overline{\mathbf{B}} &= (\Delta \mathbf{A})^{-1} (\exp(\Delta \mathbf{A}) - \mathbf{I}) \cdot \Delta \mathbf{B}.
\end{aligned}
\end{equation}

The discrete state evolution is then expressed as:
\begin{equation}
\begin{aligned}
h_t &= \overline{\mathbf{A}} h_{t-1} + \overline{\mathbf{B}} x_t, \\
y_t &= \mathbf{C} h_t.
\end{aligned}
\end{equation}

Finally, the models compute the output through a global convolution:
\begin{equation}
\begin{aligned}
\mathbf{\overline{K}} &= (\mathbf{C} \overline{\mathbf{B}}, \mathbf{C} \overline{\mathbf{A}} \overline{\mathbf{B}}, \dots, \mathbf{C} \overline{\mathbf{A}}^{M-1} \overline{\mathbf{B}}), \\
\mathbf{y} &= \mathbf{x} * \mathbf{\overline{K}},
\end{aligned}
\end{equation}
where $M$ is the input sequence length and $\overline{\mathbf{K}} \in \mathbb{R}^{M}$ is a structured convolutional kernel. This process enables efficient sequence modeling through state-space representations.

\subsection{Overall Architecture}

Figure~\ref{Fig:framework} illustrates the architecture of our LMDepth, which consists of four main modules: the Image Encoder, the Projection Module (Proj), the Modified Pyramid Spatial Pooling (MPSP), and the Depth Mamba Block (DMB).
First, the input RGB image is processed by a MobileNetV2\cite{mobilenetv2} encoder to extract multi-scale features, which are then reduced to a fixed number of channels through the Projection Module.
Second, the MPSP captures multi-scale spatial context using pooling and convolution operations. It produces three key outputs:
1) Scene classification, which provides contextual information about the input image.
2) Bins, representing discretized depth intervals for probabilistic depth estimation.
3) Global depth features, which encode global depth information and are passed to the DMB blocks via the Projection Module.
Next, each DMB combines the upscaled depth features from the previous DMB with the projected features from the corresponding encoder level. Finally, the decoder (Multiple DMB blocks) generates a probability map, where each pixel represents the likelihood of belonging to different depth bins. By performing element-wise multiplication between the probability map and the bins, LMDepth predicts a high-quality depth map from the single RGB image.

\subsection{Depth Mamba Block (DMB)}

To enhance the receptive field and strengthen visual information guidance, we propose the Depth Mamba Block (DMB), which effectively integrates depth and visual features to overcome existing limitations. The DMB takes depth features from the previous layer and visual features generated by the MobileNet encoder as inputs. By leveraging the VisionMamba (Vmamba) module~\cite{mambavision}, it achieves spatially long-range feature fusion, ensuring robust and comprehensive feature representation. The processing within the DMB is conducted as follows:

{\noindent\bf{Reshaping Features}}: Both the depth features and visual features are reshaped from a tensor of shape \((24,H, W)\) to \((H \times W, 24)\). This transformation converts the features into tokens \(H \times W\), each with a dimension of 24.

\begin{figure}[t]  
\centering
\includegraphics[width=1\linewidth]{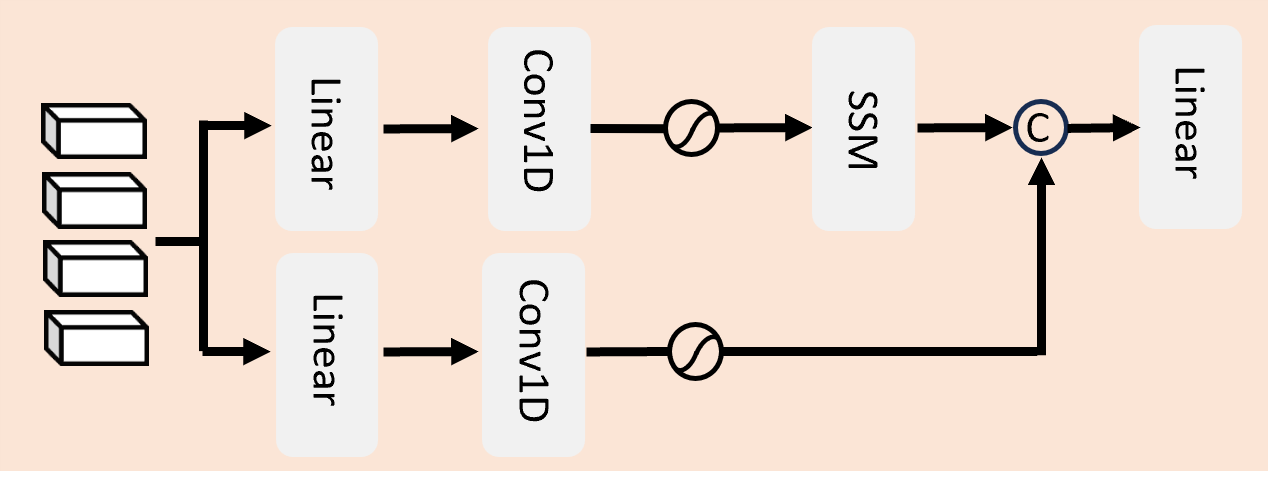}
\caption{\textbf{The overview of VMamba Module}. A one-dimensional convolution operation is added to the original Mamba to enhance the extraction of local attention.}
\label{fig:pst}
\vspace{-0.3cm}
\end{figure}
 
{\noindent\bf{Feature Fusion}}: The reshaped depth features are enhanced through the Vision Mamba (Vmamba) module, fused with visual features via another Vmamba module, and then refined to produce the final depth representation. The Vmamba module, as shown in Figure 3, is motivated by \cite{mambavision} and designed to efficiently capture global context with linear computational complexity. It processes \(h \times w\) tokens, each with a feature dimension of 24. The input tokens first pass through a normalization layer to standardize feature representation. And then, followed by a one-dimensional convolution layer which is designed to enhance Mamba's ability to extract local features. Subsequently, the SSM, as introduced in the preliminaries, extracts global spatial dependencies. The output of the SSM is combined with the original input features through a residual skip connection, ensuring the retention of essential information. The resulting features are then normalized again and fed into a Feed-Forward Network (FNN), which refines and transforms the features. A second residual skip connection is applied to the FNN output to enhance feature propagation and maintain stability during training. This structure allows Vmamba to balance computational efficiency and feature enhancement, making it an integral component of the Depth Mamba Block (DMB) for accurate and efficient depth estimation.

{\noindent\bf{Reshaping and Upsampling}}: The fused feature map is reshaped back to its original spatial dimensions and undergoes upsampling. The upsampled features are then passed to the next layer, ensuring the progressive refinement of depth estimation.


By incorporating multiple DMB, LMDepth gains an expanded receptive field and improved feature guidance, enabling more reliable depth map predictions.





\subsection{Modified Pyramid Spatial Pooling (MPSP)}
The MPSP which is modified from PSP\cite{psp}, as shown in the figure, processes the final feature map from the MobileNet\cite{mobilenetv2} encoder, denoted as \(\mathbf{F}_{\text{input}} \in \mathbb{R}^{H \times W \times C}\), and generates three outputs: scene classification, depth distribution bins, and global features. The MPSP combines spatial pooling, multi-scale feature refinement, and feature fusion to effectively extract both global context and fine-grained information.

First, the input feature map \(\mathbf{F}_{\text{input}}\) undergoes multi-scale spatial pooling, where different scales \(s \in \{1, 2, 3, 6\}\) are applied. This results in pooled feature maps \(\mathbf{P}_s \in \mathbb{R}^{\frac{H}{s} \times \frac{W}{s} \times C}\). Each pooled feature map is then processed by convolutional layers to extract refined features:
\[
\mathbf{F}_s = \text{Conv}(\mathbf{P}_s), \quad s \in \{1, 2, 3, 6\}.
\]

The refined features are then upsampled back to the original resolution:
\[
\mathbf{F}_s^{\text{up}} = \text{Upsample}(\mathbf{F}_s), \quad s \in \{1, 2, 3, 6\}.
\]

These upsampled features are concatenated with the original input feature map to form the fused feature representation:
\[
\mathbf{F}_{\text{fused}} = \text{Concat}(\mathbf{F}_{\text{input}}, \mathbf{F}_1^{\text{up}}, \mathbf{F}_2^{\text{up}}, \mathbf{F}_3^{\text{up}}, \mathbf{F}_6^{\text{up}}).
\]

From \(\mathbf{F}_{\text{fused}}\), the MPSP produces three outputs:   

{\noindent\bf{Scene Classification}}: A Multi-Layer Perceptron (MLP) is applied to the fused features to generate scene classification predictions:
\[
\mathbf{Classification} = \text{MLP}(\mathbf{F}_{\text{fused}}).
\]

{\noindent\bf{Depth Distribution Bins}}: Another MLP is used to estimate the depth distribution bins:
\[
\mathbf{Bins} = \text{MLP}(\text{Pooling}(\mathbf{F}_{\text{fused}})).
\]

{\noindent\bf{Global Features}}: A convolutional layer extracts global features from the fused representation:
\[
\mathbf{F}_{\text{global}} = \text{Conv}(\mathbf{F}_{\text{fused}}).
\]

The scene classification \(\mathbf{C}\) provides contextual information for the input image, the depth distribution bins \(\mathbf{B}\) discretize the depth range for depth estimation, and the global features \(\mathbf{G}\) encode the overall scene context. This design ensures that the MPSP head effectively combines multi-scale spatial and semantic information for robust depth estimation and scene understanding.

\begin{table*}[t]
    \centering
    \caption{Comparison with other representative lightweight depth estimation methods on the NYUDv2 dataset. The best results are highlighted in \textbf{bold}. FLOPs are computed for a single forward pass. Our models, \textbf{LMDepth} and \textbf{LMDepth-S}, achieve the lowest parameter count and FLOPs under both high-resolution and low-resolution settings, while still maintaining an obvious advantage across all evaluation metrics.}
    \label{tab:results-nyu}
    \small
    \setlength{\tabcolsep}{4.5pt}
    \renewcommand{\arraystretch}{1.2}
    \resizebox{0.98\textwidth}{!}{
    \begin{tabular}{lcccccccccc}
        \toprule
        \textbf{Method} & \textbf{Resolution} & \textbf{Architectures} & \textbf{Params (M)} & \textbf{FLOPs (G)} & $\boldsymbol{\delta_1 \uparrow}$ & $\boldsymbol{\delta_2 \uparrow}$ & $\boldsymbol{\delta_3 \uparrow}$ & \textbf{REL $\downarrow$} & \textbf{log10 $\downarrow$} & \textbf{RMS $\downarrow$} \\
        \midrule
        \rowcolor{lightgray}
        \multicolumn{11}{c}{\textbf{Low Resolution }} \\
        FastDepth~\cite{fastdepth} & 224$\times$224 & CNN & 3.9 & 0.75 & 0.771 & 0.939 & 0.976 & 0.135 & - & 0.580 \\    
        JointDepth~\cite{joint} & 240$\times$320 & CNN & 3.0 & 1.38 & 0.778 & 0.942 & 0.987 & 0.142 & - & 0.591 \\
        TuMDE~\cite{tumde} & 240$\times$320 & CNN & 4.6 & 0.95 & 0.772 & 0.948 & 0.987 & 0.162 & 0.067 & 0.536 \\
        YinMDE~\cite{enforcing} & 338$\times$338 & CNN & 2.7 & 5.60 & 0.813 & 0.958 & 0.991 & 0.135 & 0.060 & - \\
        DaNet~\cite{danet} & 228$\times$304 & Transformer & 8.2 & 1.50 & 0.829 & 0.963 & 0.991 & 0.140 & 0.059 & 0.487 \\
        GuideDepth-S~\cite{guidedepth} & 240$\times$320 & CNN & 5.7 & 1.52 & 0.812 & 0.958 & 0.989 & 0.144 & 0.060 & 0.514 \\
        GuideDepth~\cite{guidedepth} & 240$\times$320 & CNN & 5.8 & 2.63 & 0.823 & 0.961 & 0.990 & 0.138 & 0.058 & 0.501 \\
        \textbf{LMDepth-S (Ours)} & 240$\times$320 & Mamba & 2.3 & 0.59 & 0.825 & 0.965 & 0.991 & 0.140 & 0.059 & 0.476 \\
        \textbf{LMDepth (Ours)} & 240$\times$320 & Mamba & 2.9 & 0.72 & \textbf{0.833} & \textbf{0.968} & \textbf{0.993} & \textbf{0.134} & \textbf{0.055} & \textbf{0.471} \\
        \midrule
        \rowcolor{lightgray}
        \multicolumn{11}{c}{\textbf{High Resolution}} \\
        FastDepth~\cite{fastdepth} & 480$\times$640 & CNN & 3.9 & 4.00 & 0.778 & 0.942 & 0.987 & 0.142 & - & 0.591 \\
        JointDepth~\cite{joint} & 480$\times$640 & CNN & 3.0 & 6.49 & 0.790 & 0.955 & 0.990 & 0.149 & - & 0.565 \\
        TuMDE~\cite{tumde} &480$\times$640&CNN&4.6 & 4.60 & 0.772 & 0.948 & 0.987 & 0.162 & 0.067 & 0.536 \\
        YinMDE~\cite{enforcing} &480$\times$640&CNN&2.7 & 15.80 & 0.813 & 0.958 & 0.991 & 0.135 & 0.060 & - \\
        DaNet~\cite{danet} &480$\times$640&Transformer&8.2 & 5.51 & 0.829 & 0.963 & 0.991 & 0.140 & 0.059 & 0.487 \\
        GuideDepth-S~\cite{guidedepth} & 480$\times$640 & CNN & 5.7 & 6.03 & 0.830 & 0.969 & 0.991 & 0.131 & 0.057 & 0.491 \\
        GuideDepth~\cite{guidedepth} & 480$\times$640 & CNN & 5.8 & 10.47 & 0.844 & 0.971 & 0.992 & 0.129 & 0.055 & 0.449 \\
        \textbf{LMDepth-S (Ours)} & 480$\times$640 & Mamba & 2.3 & 2.10 & 0.833 & 0.970 & 0.993 & 0.135 & 0.057 & 0.463 \\
        \textbf{LMDepth (Ours)} & 480$\times$640 & Mamba & 2.9 & 2.77 & \textbf{0.854} & \textbf{0.977} & \textbf{0.995} & \textbf{0.123} & \textbf{0.052} & \textbf{0.438} \\
        \bottomrule
    \end{tabular}
    }
\end{table*}

\begin{table*}[t]
    \centering
    \caption{Comparison with other representative lightweight depth estimation methods on the KITTI datasets. The best results are highlighted in \textbf{bold}. FLOPs are computed for a single forward pass. Our models, \textbf{LMDepth} and \textbf{LMDepth-S}, achieve the lowest parameter count and FLOPs under both high-resolution and low-resolution settings, while still maintaining an obvious advantage across all evaluation metrics.}
    \label{tab:results-kitti}
    \small
    \renewcommand{\arraystretch}{1.2} 
    \setlength{\tabcolsep}{4.5pt} 
    \begin{tabular}{lcccccccccc}
        \toprule
        \textbf{Method} & \textbf{Resolution} & \textbf{Architectures} & \textbf{Params (M)} & \textbf{FLOPs (G)} & $\boldsymbol{\delta_1 \uparrow}$ & $\boldsymbol{\delta_2 \uparrow}$ & $\boldsymbol{\delta_3 \uparrow}$ & \textbf{REL $\downarrow$} & \textbf{Sq-rel $\downarrow$} & \textbf{RMSE $\downarrow$} \\
        \midrule
        \rowcolor{lightgray}
        \multicolumn{11}{c}{\textbf{Low Resolution }} \\
        FastDepth~\cite{fastdepth} & 192$\times$640 & CNN & 3.9 & 1.82 & 0.778 & 0.942 & 0.987 & 0.142 & 0.591 & - \\
        TuMDE~\cite{tumde} & 192$\times$640 & CNN & 4.6 & 2.19 & 0.760 & 0.930 & 0.980 & 0.150 & - & 5.80 \\
        GuideDepth~\cite{guidedepth} & 192$\times$640 & CNN & 5.8 & 4.20 & 0.868 & 0.968 & 0.991 & 0.114 & - & 4.15 \\
        DaNet~\cite{danet} &192$\times$620&Transformer&8.2 & 3.50 & 0.885 & 0.972 & 0.990 & 0.110 & 0.550 & 6.20 \\
        LMDepth-S (Ours) & 192$\times$620 & Mamba & 2.3 & 0.82 & 0.903 & 0.982 & 0.996 & 0.092 & 0.463 & 3.71 \\
        LMDepth (Ours) & 192$\times$620 & Mamba & 2.9 & 1.08 & \textbf{0.908} & \textbf{0.983} & \textbf{0.996} & \textbf{0.089} & \textbf{0.441} & \textbf{3.63} \\
        \midrule
        \rowcolor{lightgray}
        \multicolumn{11}{c}{\textbf{High Resolution }} \\
        FastDepth~\cite{fastdepth} & 384$\times$1260 & CNN & 3.9 & 6.17 & 0.825 & 0.961 & 0.991 & 0.136 & 0.714 & 4.33 \\
        TuMDE~\cite{tumde} &384$\times$1260&CNN&4.6 & 4.50 & 0.770 & 0.910 & 0.980 & 0.145 & 0.630 & 6.50 \\
        GuideDepth~\cite{guidedepth} & 384$\times$1260 & CNN & 5.8 & 16.75 & 0.882 & 0.972 & 0.991 & 0.109 & - & 4.10 \\
        DaNet~\cite{danet} & 384$\times$1260 & Transformer & 8.2 & 7.85 & 0.920 & 0.981 & 0.995 & 0.086 & 0.411 & 3.52 \\
        LMDepth-S & 384$\times$1260 & Mamba & 2.3 & 3.21 & 0.925 & 0.987 & 0.997 & 0.080 & 0.357 & 3.31 \\
        LMDepth & 384$\times$1260 & Mamba & 2.9 & 4.05 & \textbf{0.926} & \textbf{0.988} & \textbf{0.997} & \textbf{0.079} & \textbf{0.352} & \textbf{3.28} \\
        \bottomrule
    \end{tabular}
\end{table*}

\subsection{Loss Functions}

Auxiliary scene classification is an auxiliary subtask designed to provide implicit guidance for the generation of depth distribution bins. The top-level Modified PSPHead Module extracts global semantic information using convolutional neural networks (CNNs) and Multi-Layer Perceptrons (MLPs). A three-layer prediction module with a SiLU activation function is applied to the output to produce the final classification results. During training, a simple cross-entropy classification loss $\mathcal{L}_{cls}$ is used, which is defined as:

\begin{equation}
\mathcal{L}_{cls}(\mathbf{y}, \hat{\mathbf{y}}) = \log(\hat{\mathbf{y_k}})
\label{eq:L_pixel}
\end{equation}

where $\hat{\mathbf{y}}$ represents the predicted probabilities for each scene, and $\mathbf{y}$ represents the ground truth.

After predicting the final depth maps, we adopt a scaled version of the Scale-Invariant loss (SI) introduced by Eigen~\cite{eigen2014} as the depth regression loss. The loss is formulated as follows:

\begin{equation}
\mathcal{L}_{reg} = \alpha \sqrt{\frac{1}{T}\sum_i g_{i}^{2} - \frac{\lambda}{T^2}(\sum_i g_i)^2},
\label{eq::loss}
\end{equation}

\noindent where $g_i = \log \Tilde{d_i} - \log d_i$, with $\Tilde{d_i}$ being the predicted depth value and $d_i$ being the ground truth depth value. $T$ denotes the number of pixels with valid ground truth depth values. Following~\cite{AdaBins}, we set $\lambda = 0.85$ and $\alpha = 10$ for all experiments.

The final total loss function is a weighted sum of the two losses. The total loss is expressed as:

\begin{equation}
\mathcal{L}_{total} = \mathcal{L}_{cls} + \beta \cdot \mathcal{L}_{reg}.
\end{equation}

\section{Experiments}
 In this section, we evaluate the performance of LMDepth by comparing it with several lightweight networks. First, we introduce the datasets, evaluation metrics, and implementation details. Then, present quantitative comparisons to the other lightweight networks in supervised monocular depth estimation.
\subsection{Experimental Settings}

{\noindent\bf{Datasets.}} To evaluate outdoor and indoor scenarios, we evaluate our methods on the NYUDv2 and KITTI datasets. NYUDv2 is an indoor dataset that collects 464 scenes with 120K pairs of RGB and depth maps. KITTI is a dataset that provides stereo images and corresponding 3D laser scans of outdoor scenes captured by equipment mounted on a moving vehicle.

{\noindent\bf{Baseline.}} For the fair comparison, we introduce different representative lightweight depth estimation methods, including 1) methods with CNN-base network architecture (e.g., FastDepth~\cite{fastdepth}, TuMDE~\cite{tumde} YinMDE~\cite{enforcing}, GuideDepth~\cite{guidedepth}, JointDepth~\cite{joint}). 2) methods with Transformer-based architecture (e.g., DaNet~\cite{danet}. Different from these methods,
We provide two Mamba-based models: LMDepth and LMDepth-S. The LMDepth-S is the more lightweight version of LMDepth, reducing complexity with the MPSP pooling scales from \((1, 2, 3, 6)\) to \((1, 6)\) and the feature projection dimensions in the decoder from 24 to 16.

{\noindent\bf{Evaluation metrics.}} We use the standard six metrics used in prior works on depth estimation  \cite{eigen2014} to compare our method against other methods. These error metrics are defined as:
average relative error (REL): $\frac{1}{n}\sum_p^n \frac{\lvert y_p-\hat{y}_p \rvert}{y}$;
root mean squared error (RMS): $\sqrt{\frac{1}{n}\sum_p^n (y_p-\hat{y}_p)^2)}$;
average ($\log_{10}$) error: $\frac{1}{n}\sum_p^n \lvert \log_{10}(y_p)-\log_{10}(\hat{y}_p) \rvert$;
threshold accuracy ($\delta_i$): $\%$ of $y_p$ s.t. $\text{max}(\frac{y_p}{\hat{y}_p},\frac{\hat{y}_p}{y_p}) = \delta < thr$ for $thr=1.25,1.25^2,1.25^3$;
where $y_p$ is a pixel in the depth image $y$, $\hat{y}_p$ is a pixel in the predicted depth image $\hat{y}$, and $n$ is the total number of pixels for each depth image. 
Additionally for KITTI, we use the two standard metrics: Squared Relative Difference (Sq.~Rel): $\frac{1}{n}\sum_p^n \frac{\|y_p-\hat{y}_p \|^2}{y}$; 
and RMSE log: $\sqrt{\frac{1}{n}\sum_p^n \|\log y_p - \log \hat{y}_p\|^2}$.

{\noindent\bf{Implementation Details.}}
We implement the proposed network in PyTorch\cite{pytorch}  and trained on 4 $\times$ NVIDIA 2080Ti 12GB GPU. For model training, we use the AdamW\cite{adam} optimizer with weight-decay $10^{-4}$ and set the hyperparameter \(\beta\) to 0.1. We train our model for 20 epochs with a batch size of 12, and the initial learning rate is set to 0.0002 and reduced by 10 $\%$ for every 10 epochs. For the NYUDepth-v2 dataset, we utilize the official 25 classes divided by folder names for the auxiliary scene understanding task. For KITTI, since the outdoor dataset is tough to classify, we omit the scene classification loss and only use ground truth depth to provide supervision.

\subsection{Main Results}

Table~\ref{tab:results-nyu} compares the performance and computational efficiency of different lightweight depth estimation methods on the indoor dataset NYUDv2~\cite{nyu}. It highlights the ability of our approach, LMDepth, to achieve high-accuracy depth reconstruction while maintaining low computational overhead across both low and high-resolution settings. Several key observations can be summarized: (1) Our LMDepth model achieves the best performance across all evaluation metrics, including \(\delta_1\), \(\delta_2\), and RMS. At low resolution (\(240 \times 320\)), LMDepth reaches \(\delta_1 = 0.833\) and REL = 0.134, while at high resolution (\(480 \times 640\)), it further improves to \(\delta_1 = 0.854\) and REL = 0.123, outperforming other methods. (2) LMDepth significantly reduces FLOPs while maintaining a low parameter count. For low-resolution inputs, the Base model requires only 0.72 GFLOPs, while the lightweight -S model further reduces this to 0.59 GFLOPs with minimal accuracy loss. At high resolution, the Base model achieves superior performance with just 2.77 GFLOPs, far lower than methods like GuideDepth (5.72 GFLOPs). The overall efficiency demonstrates that LMDepth offers a superior trade-off between computational complexity and depth accuracy, which is crucial for real-time deployment on embedded systems.

Table~\ref{tab:results-kitti} presents a comparison of various lightweight depth estimation methods on the outdoor dataset KITTI~\cite{kitti}, highlighting the performance and computational efficiency of our LMDepth approach. Our method consistently achieves the best results across all evaluation metrics. At low resolution (\(192 \times 620\)), LMDepth reaches \(\delta_1 = 0.908\) and REL = 0.089, outpacing other models. At high resolution (\(384 \times 1260\)), LMDepth further improves these results with \(\delta_1 = 0.926\) and REL = 0.079, clearly outperforming competitors in both accuracy and efficiency. In terms of computational efficiency, LMDepth requires significantly fewer FLOPs than competing methods. For low-resolution inputs, the Base model consumes only 1.08 GFLOPs, while the lightweight -S version further reduces this to 0.82 GFLOPs with only a minimal loss in accuracy. At high resolution, the Base model achieves superior performance with just 4.05 GFLOPs, well below methods such as GuideDepth, which requires 16.75 GFLOPs. This combination of high accuracy and low computational overhead makes LMDepth an optimal choice for real-world deployment.

Figure~\ref{fig:NYUDv2_qualitative} showcases qualitative depth estimation results from various methods, including FastDepth~\cite{fastdepth}, TuMDE~\cite{tumde}, GuideDepth~\cite{guidedepth}, and our approach. The results are evaluated across both indoor and outdoor scenarios from the NYUDv2 dataset. As depicted in the figure, our method consistently delivers more accurate and visually consistent depth estimations, with predictions that closely match the ground truth in a wide range of real-world environments. In particular, our approach excels in handling challenging scenarios such as low-texture areas and complex indoor environments, where other methods like FastDepth and TuMDE struggle to maintain accuracy. Additionally, the superior performance of our method in outdoor scenes, with its ability to capture fine-grained depth variations, highlights its robustness across diverse settings. The visual comparisons demonstrate that our method not only improves the overall depth accuracy but also preserves fine details in the depth map, particularly in regions with subtle depth transitions, where other models fail to capture such intricate variations. This provides strong evidence of the effectiveness and generalizability of our approach for practical depth estimation tasks.

\begin{figure}[t]  
\centering
\includegraphics[width=1.0\linewidth]{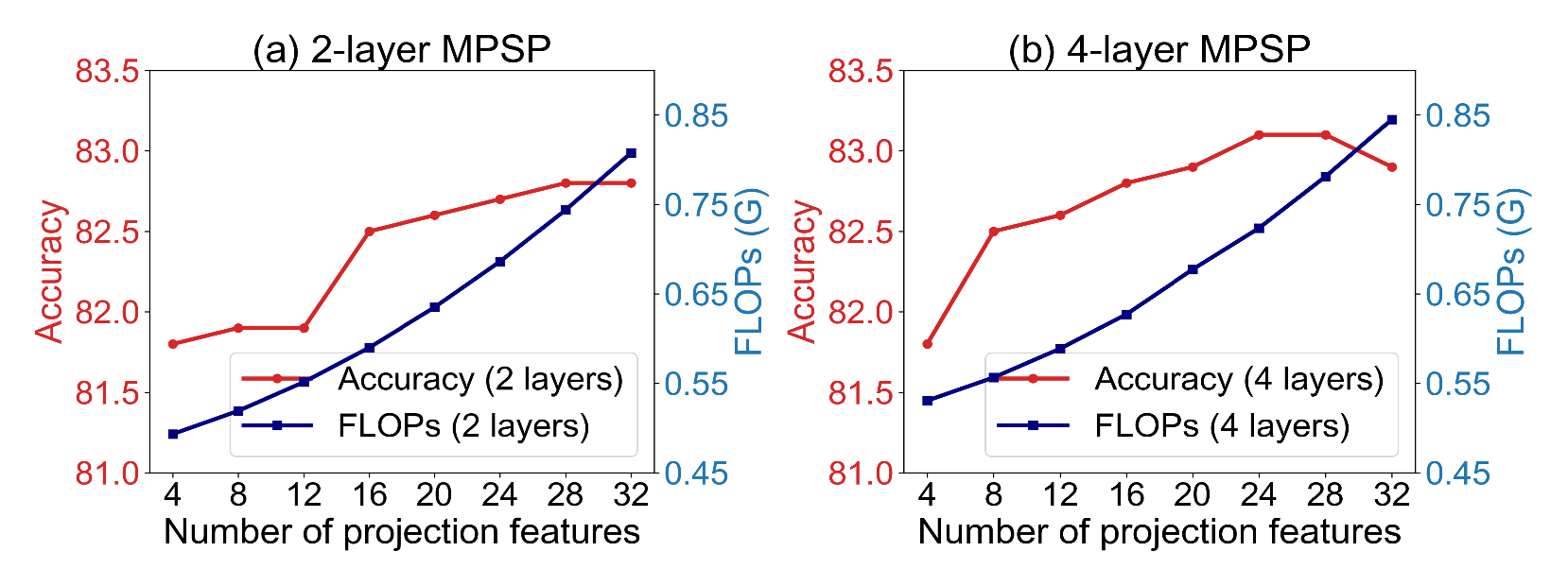}
\caption{\textbf{Varing projection features on 2 and 4 MPSP layers.} \textcolor{red}{Red lines} show the trend of \(\delta_1\) accuracy, and \textcolor{blue}{Blue lines} show trend of FLOPs. LMDepth is selected for the highest accuracy, and LMDepth-S is selected for the equilibrium point between efficiency and performance. }
\label{Fig:ablation on layers}
\end{figure}

 \begin{figure*}[t]
\vspace{0.1cm} 
\centering
\includegraphics[width=1\linewidth]{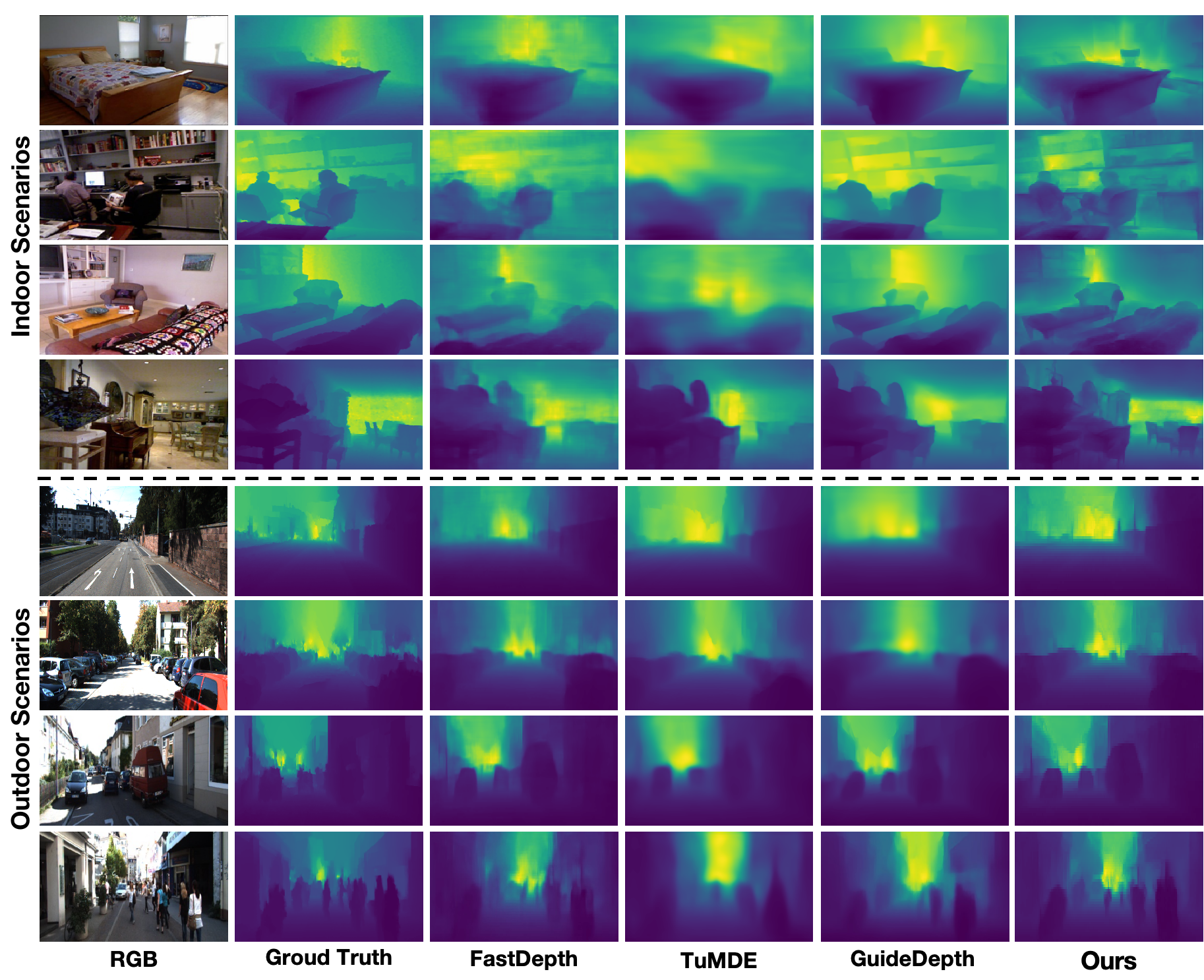}
\caption{\textbf{Qualitative comparison of various lightweight depth estimation methods on the indoor and outdoor scenarios.} Our predictions are more refined and closer to ground truth. It demonstrates that our method achieves better depth estimation of distant objects and finer details.}
 \label{fig:NYUDv2_qualitative}
\end{figure*}

\subsection{Ablation Studies}
Table~\ref{table:decoder-results} evaluates the effectiveness of the proposed VMamba module in the decoder's Depth Mamba Block (DMB) by replacing it with transformer-based and CNN-based decoder. The results highlight the superiority of VMamba in terms of both accuracy and computational efficiency across different datasets.
For example, 
on the NYUDv2 dataset, VMamba achieves the best performance with a \(\delta_1\) accuracy of 0.830 and an RMS error of 0.472. Notably, the parameter count of VMamba (2.9M) matches that of the CNN-based decoder, and its FLOPs (0.73G) are only marginally higher than CNN's 0.69G, yet significantly lower than the Transformer-based decoder, which requires 2.8G FLOPs. This demonstrates that VMamba can capture global context more effectively than CNN while maintaining a computational cost far lower than Transformer, which suffers from quadratic complexity due to its attention mechanism.
These findings validate the effectiveness of VMamba in balancing accuracy and efficiency, making it a robust choice for lightweight monocular depth estimation across both indoor and outdoor datasets.

\begin{figure*}[t]  
\centering
\includegraphics[width=1\linewidth]{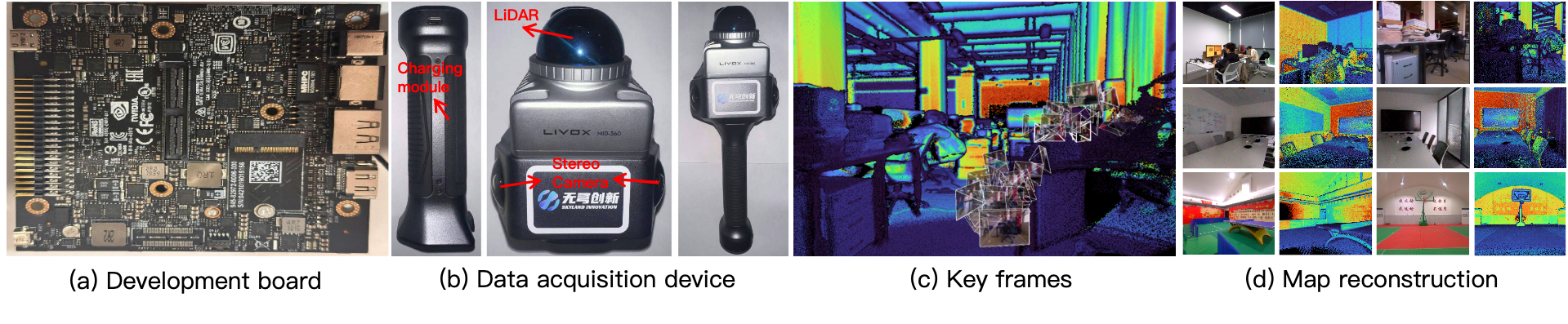}
\vspace{-0.3in}
\caption{\textbf{Embedded deployment platform and data acquisition pipeline.} (a) The lightweight deployment platform is based on the NVIDIA Jetson AGX Xavier, equipped with a Volta GPU featuring 512 CUDA cores and 64 Tensor cores.
(b) The mobile data acquisition device comprises a LiDAR scanner (Livox MID360) and a smart handheld controller for real-time control and synchronization.
(c) The dense mapping process is built upon multimodal keyframe registration. During motion, the system captures synchronized RGB and depth images, aligns them with LiDAR point clouds, and constructs a globally consistent 3D map. Camera icons represent keyframe poses, and color-coded point clouds indicate relative depth.
(d) Sample reconstruction results from our collected indoor dataset, covering diverse scenes such as offices, meeting rooms, and sports spaces.}
\label{Fig:chart}
\end{figure*}

Figure~\ref{Fig:chart} compares the accuracy and FLOPs for different numbers of projection features with 2-layer and 4-layer pooling in MPSP. 
It reveals that as the projection features increase (represented by the blue line), FLOPs also increase. While the accuracy (represented by the red line) initially improves with increasing complexity, it eventually plateaus and even declines.
We select the model with the highest accuracy as LMDepth, and select the model with a point of maximum gradient as LMDepth-S.

\begin{table}[t]
    \centering
    \caption{Ablation stueis of VMamba module. Compared with CNN and Transformer, Mamba has similar parameters and FLOPs to CNN while achieving slightly higher accuracy than Transformer.}
    \label{table:decoder-results}
    \small
    \resizebox{0.48\textwidth}{!}{
    \begin{tabular}{lccccc}
        \toprule
        \textbf{Dataset} & \textbf{Decoders} & \textbf{Parameters} & \textbf{FLOPs} & \textbf{RMS $\downarrow$} & $\delta_1 \uparrow$ \\
        \midrule
         & Mamba       & 2.9M & 0.73G & 0.472 & 0.833 \\
        NYUDv2    & Transformer & 5.4M & 2.8G  & 0.473 & 0.828 \\
                                           & CNN         & 3.1M & 0.69G & 0.477 & 0.822 \\
        \midrule
          & Mamba       & 2.9M & 1.08G & 3.63  & 0.908 \\
         KITTI  & Transformer & 5.4M & 2.8G  & 3.65  & 0.904 \\
                        & CNN         & 3.1M & 1.08G & 3.64  & 0.904 \\
        \bottomrule
    \end{tabular}
    }
\end{table}

\section{Lightweight Real-World Deployment}
In this section, we present a practical deployment strategy for the LMDepth model. We first evaluate its real-time inference performance on the target embedded platform to ensure it meets the dual requirements of accuracy and efficiency in real-world applications. Furthermore, we conduct generalization tests across diverse real-world environments to verify the model's robustness and adaptability under varying scene conditions.

\subsection{Deployment setting}
To enable real-world deployment, we first convert the trained PyTorch model into ONNX format, followed by INT8 static quantization for improved inference speed and model compactness. The optimized model is then deployed on the NVIDIA Jetson AGX Xavier platform, which supports efficient execution through TensorRT. To evaluate system performance under practical conditions, we conduct experiments using our custom mobile data acquisition device, capturing RGB images and LiDAR point clouds across diverse indoor scenes. The inference results, tested directly on the embedded platform, validate both the speed and robustness of the deployed model in real-world settings. The embedding device is shown in Figure~\ref{Fig:chart} (a).

{\noindent\bf{Model Quantization for Efficient Deployment}}. To enable efficient edge deployment, we convert the trained PyTorch model to ONNX format and apply post-training static quantization. The ONNX format allows for lightweight cross-platform inference with graph-level optimizations such as operator fusion and constant folding. Static quantization further compresses the model by converting 32-bit floating-point weights and activations to 8-bit integers while preserving numerical fidelity through calibrated scaling and zero-point encoding. This quantization process significantly reduces model size and improves inference speed, particularly when combined with TensorRT acceleration on the embedded platform. In practice, our quantized model achieves over 4× compression and nearly 2× speed-up with minimal accuracy loss, making it well-suited for real-time applications.

{\noindent\bf{Embedded Inference Platform}}. We deploy our model on the Jetson AGX Xavier platform, which features 512 CUDA cores and 64 Tensor cores, offering powerful embedded AI inference capabilities. The platform supports INT8 quantization and is compatible with various mainstream deep learning frameworks, as well as a wide range of peripheral interfaces, making it well-suited for edge deployment scenarios. The key frame selection and depth map reconstruction are illustrated in Figure~\ref{Fig:chart} (b).

{\noindent\bf{Data Acquisiotion Device}}. We use a custom-built sensor system based on the Livox MID-360, capable of simultaneously capturing synchronized point clouds and RGB images. To improve data efficiency, frames with significant pose variation are selected as keyframes. For each keyframe, the corresponding point cloud is aligned with the RGB image using calibrated camera parameters, and 3D points are projected to generate depth maps. This process ensures spatially consistent depth images for downstream model training. The key frame selection and depth map reconstruction are illustrated in Figure~\ref{Fig:chart} (c) and (d).

\subsection{Experiments}
Table~\ref{table:ownDataset} shows the generalization tests of different lightweight depth estimation methods (i.e., Fastdepth, TuMDE, GuideDepth, DaNet) on our self-collected
datasets. The experimental results demonstrate the superior performance of our proposed method in terms of accuracy. Specifically, the $\delta_1$ accuracy reaches 0.54, showing a significant improvement over baseline methods. The relative error (REL) is reduced to 0.24, which is notably better than the state-of-the-art lightweight solution GuideDepth~\cite{guidedepth} (REL = 0.26). These substantial improvements in quantitative metrics strongly validate the generalization capability of our approach.  In terms of real-time performance, comparative experiments further verify the effectiveness of our method. The pruned LMDepth-T model achieves an inference speed of 120 FPS on our experimental platform. Although this is slightly lower than FastDepth~\cite{fastdepth} (174 FPS) and GuideDepth (170 FPS), it significantly surpasses the baseline requirement of 24 FPS for real-time systems. Furthermore, the inference speed of LMDepth-T is markedly higher than DaNet~\cite{danet}, which is based on a Transformer~\cite{transformer} backbone, while delivering comparable accuracy. This result highlights the contribution of Mamba~\cite{mamba} in improving inference efficiency. Overall, the proposed solution achieves a favorable balance between accuracy and speed, offering sufficient temporal margins for downstream image processing tasks. Such performance advantages provide a solid foundation for real-world deployment in complex scenarios with real-time constraints.

\begin{figure}[t]  
\centering
\includegraphics[width=1\linewidth]{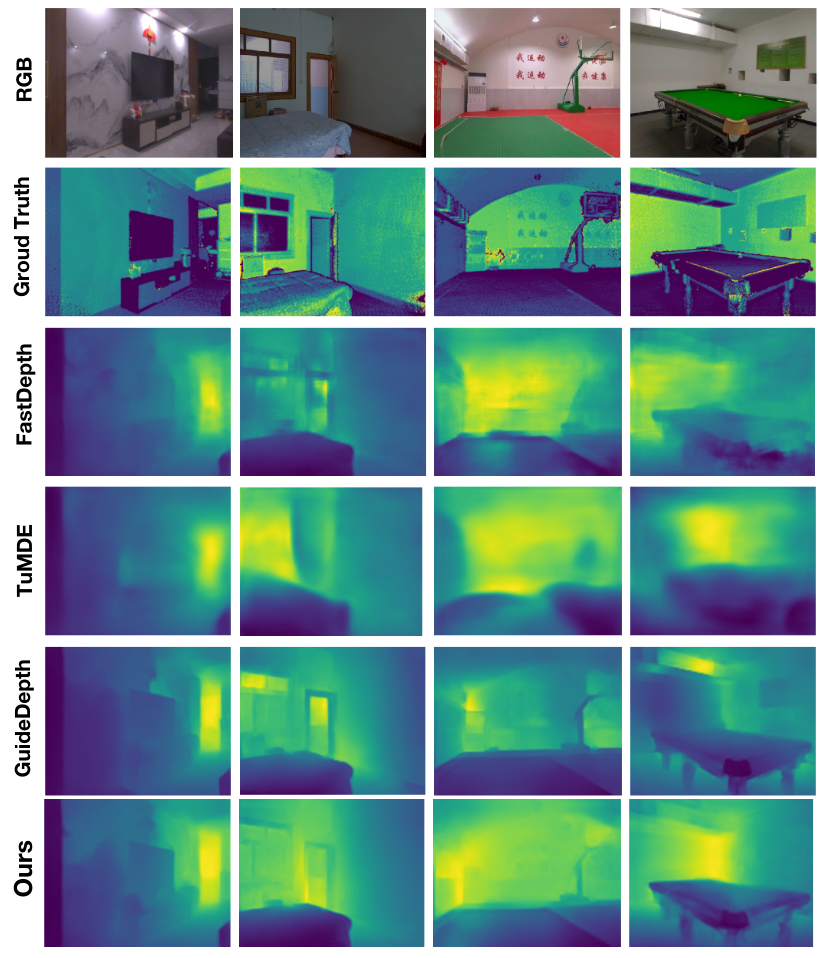}
\caption{\textbf{Qualitative comparison of various lightweight depth estimation methods on real-world scenarios.} Our predictions are more refined and closer to ground truth. It demonstrates that our method achieves better depth estimation of distant objects and finer details.}
\label{Fig:real-world-vis}
\end{figure}

Figure~\ref{Fig:real-world-vis} shows the qualitative comparison of various lightweight depth estimation
methods (i.e., Fastdepth, TuMDE, GuideDepth) on real-world scenarios. All models are trained on the NYUDv2~\cite{nyu} dataset and evaluated across both seen scenes (e.g., bedrooms, living rooms) and unseen scenes (e.g., basketball courts, billiard halls). As shown in the first two columns, our method yields accurate estimations with well-preserved object boundaries. For instance, in the bedroom scene, our model clearly delineates the bed and door frame, where other methods struggle. GuideDepth, in particular, suffers from severe depth discontinuities around structural edges, while our predictions remain geometrically consistent.

Table \ref{tab:quantization_comparison} shows the comparison of our model performance before and after quantization. We first convert the PyTorch-trained model into ONNX format. As ONNX preserves only the inference-relevant computation graph and parameters while removing redundant operations, the model size is significantly reduced without any loss in accuracy. On the NVIDIA Jetson AGX Xavier platform, the ONNX model achieves the same accuracy as its PyTorch counterpart while reducing the model size from 26MB to 8.54MB and increasing inference speed from 62 FPS to 84 FPS (35.5\% improvement) at a resolution of 320$\times$280. Further applying INT8 quantization compresses the model to just 2.63MB—only 10.1\% of the original size—while maintaining comparable accuracy (0.84 vs. 0.86) and boosting the inference speed to 122 FPS. These results demonstrate that our lightweight model substantially enhances runtime efficiency with negligible accuracy degradation, offering a practical solution for real-time, multi-platform deployment.

\begin{table}[t]
    \centering
    \caption{Zero-shot generalization results on our self-collected dataset.}
    \label{table:ownDataset}
    \resizebox{0.5\textwidth}{!}{ 
    \begin{tabular}{lccccccc}
        \toprule
        Methods & REL $\downarrow$ & RMS $\downarrow$ & log10 $\downarrow$ & $\delta_1 \uparrow$ & $\delta_2 \uparrow$ & $\delta_3 \uparrow$ & FPS $ \uparrow$ \\
        \midrule
        Fastdepth~\cite{fastdepth} & 0.51 & 1.86 & 0.31 & 0.25 & 0.50 & 0.70 & 174 \\
        TuMDE~\cite{joint} & 0.54 & 1.96 & 0.35 & 0.21 & 0.45 & 0.60 & 160 \\
        GuideDepth~\cite{guidedepth} & 0.28 & 1.18 & 0.14 & 0.51 & 0.79 & 0.90 & 170 \\
        DaNet~\cite{danet} & 0.25 & 1.12 & 0.13 & 0.53 & 0.82 & 0.92 & 12 \\
        LMDepth (Ours) & 0.24 & 1.09 & 0.11 & 0.55 & 0.86 & 0.94 & 120 \\
        \bottomrule
    \end{tabular}
    }
\end{table}

\begin{table}[t]
    \centering
    \caption{Comparison of model performance before and after quantization. ``LR'' and ``HR'' refer to low (320×280) and high (640×480) resolution respectively.}
    \label{tab:quantization_comparison}
    \begin{tabular}{lccccc}
    \toprule
    Model & $\delta_2$ & LR (FPS) & HR (FPS) & Size (MB) \\
    \midrule
    LMDepth (.pt) & 0.86 & 62 & 20 & 8.54 \\
    LMDepth (.onnx) & 0.86 & 84 & 31 & 4.22 \\
    LMDepth (Quantized) & 0.84 & 122 & 43 & 2.63 \\
    \bottomrule
    \end{tabular}
\end{table}

\section{Conclusion}

We propose LMDepth, a lightweight monocular depth estimation network built on the Mamba-based framework, designed for high-precision depth estimation with low computational overhead. LMDepth incorporates novel components like Modified Pyramid Spatial Pooling (MPSP) and Depth Mamba Block (DMB) to efficiently extract global context and fuse image and depth features. Evaluated on benchmark datasets (e.g., NYUDv2, KITTI), LMDepth outperforms state-of-the-art lightweight methods, achieving competitive accuracy with significantly lower parameters and FLOPs. Moreover, we deploy LMDepth on an embedded platform with INT8 quantization, validating its practical value for real-world edge applications. Its cross-dataset generalization further demonstrates robustness, making it adaptable to resource-constrained platforms. This work sets a new benchmark for lightweight depth estimation and highlights the potential of Mamba for broader vision tasks.

\bibliographystyle{IEEEtran}
\bibliography{main}
\end{document}